\documentclass[sigconf, nonacm]{acmart}

\usepackage{booktabs} 
\usepackage{graphicx}
\usepackage{float}
\usepackage{indentfirst}
\usepackage{cleveref}
\usepackage[lofdepth,lotdepth]{subfig}
\usepackage{enumitem}
\usepackage[font=small,labelfont=bf]{caption}

\acmArticle{4}
\acmPrice{15.00}

\begin{document}
\title{CapStore: Energy-Efficient Design and Management of the On-Chip Memory for CapsuleNet Inference Accelerators}

\author{\textsuperscript{1}Alberto Marchisio, \textsuperscript{1}Muhammad Abdullah Hanif, \textsuperscript{2}Mohammad Taghi Teimoori,\\
and \textsuperscript{1}Muhammad Shafique}
\affiliation{\textsuperscript{1}Vienna University of Technology, Vienna, Austria \\
\textsuperscript{2}Sharif University of Technology, Teheran, Iran \\
\{alberto.marchisio,muhammad.hanif,muhammad.shafique\}@tuwien.ac.at, teimoori@ce.sharif.edu}

\renewcommand{\shortauthors}{A. Marchisio et al.}

\begin{abstract}
Deep Neural Networks (DNNs) have been established as the state-of-the-art algorithm for advanced machine learning applications. Recently, CapsuleNets have improved the generalization ability, as compared to DNNs, due to their multi-dimensional capsules. However, they pose high computational and memory requirements, which makes energy-efficient inference a challenging task. In this paper, we perform an extensive analysis to demonstrate their key limitations due to intense memory accesses and large on-chip memory requirements. To enable efficient CaspuleNet inference accelerators, we propose a specialized on-chip memory hierarchy which minimizes the off-chip memory accesses, while efficiently feeding the data to the accelerator. We analyze the on-chip memory requirements for each memory component of the architecture. By leveraging this analysis, we propose a methodology to explore different on-chip memory designs and a power-gating technique to further reduce the energy consumption, depending upon the utilization across different operations of a CapsuleNet. Our memory designs can significantly reduce the energy consumption of the on-chip memory by up to 86\%, when compared to a state-of-the-art memory design. Since the power consumption of the memory elements is the major contributor in the power breakdown of the CapsuleNet accelerator, as we will also show in our analyses, the proposed memory design can effectively reduce the overall energy consumption of the complete CapsuleNet accelerator architecture.
\end{abstract}

%
%



\maketitle


\vspace*{-2mm}
\section{Introduction}
Deep Neural Networks (DNNs) have shown promising results for various machine learning (ML)-based applications, e.g., image and video processing, automotive, medicine, and finance, but at a cost of high computational complexity, energy consumpyion, and memory requirements. To reduce the energy and latency, many researchers have designed specialized DNN inference accelerators~\cite{ref:Cnvlutin} \cite{ref:DaDianNao} \cite{ref:Eyeriss} \cite{ref:EIE} \cite{ref:TPU} \cite{ref:FlexFlow} \cite{ref:SCNN}. Recently, Sabour and Hinton et al.~\cite{ref:dyn_routing} investigated the \textit{CapsuleNets}, a particular type of DNNs which has multi-dimensional capsules instead of uni-dimensional neurons (as used in traditional DNNs). The ability to encapsulate hierarchical information of different features (position, orientation, scaling) in a single capsule allows to achieve high accuracy in computer vision applications (e.g., MNIST~\cite{ref:MNIST} handwritten digits classification).

To tackle the challenge of performing efficiently CapsuleNet inference, systolic array-based hardware accelerators can be employed for inference hardware architectures. Recently, Marchisio et al. \cite{ref:capsacc_arxiv} proposed a systolic array-based hardware accelerator to improve the performance efficiency of CapsuleNet inference, compared to GPUs. This work primarily focused on the computational parts using an array of accelerators, which optimizes the routing-by-agreement algorithm, but ignores the memory architecture design and management for such hardware accelerators, which is a crucial component when considering energy reductions of the overall hardware design. Our experimental analysis in \Cref{sec:analysis} illustrates that the memory energy for both the on-chip and off-chip contributes to 96\% of the total energy consumption. Therefore, only employing an accelerator-based processing array is not sufficient to achieve a high energy efficiency. It is crucial to to invest further effort to analyze the possibilities and opportunities to reduce the total/overall memory energy. The assumptions made by many DNN accelerator architectures (like \cite{ref:TPU} and \cite{ref:SCNN}) and the recent CapsuleNet accelerator \cite{ref:capsacc_arxiv} of having a huge on-chip memory is not applicable in embedded applications (e.g., deployed in the IoT-edge devices), where the hardware resources are constrained and memory resources are scarce. Hence, there is a significant need to investigate energy-efficient design and management of on-chip memory hierarchy for CapsuleNet hardware architectures to enable their embedded inference. To understand the memory design challenges and the optimization potential for CapsuleNet accelerator-based architectures, we perform a detailed analysis of the memory requirements in terms of size, bandwidth and accesses for every stage of the CapsuleNet inference.

\begin{figure}[t]
	\centering
	\includegraphics[width=.9\linewidth]{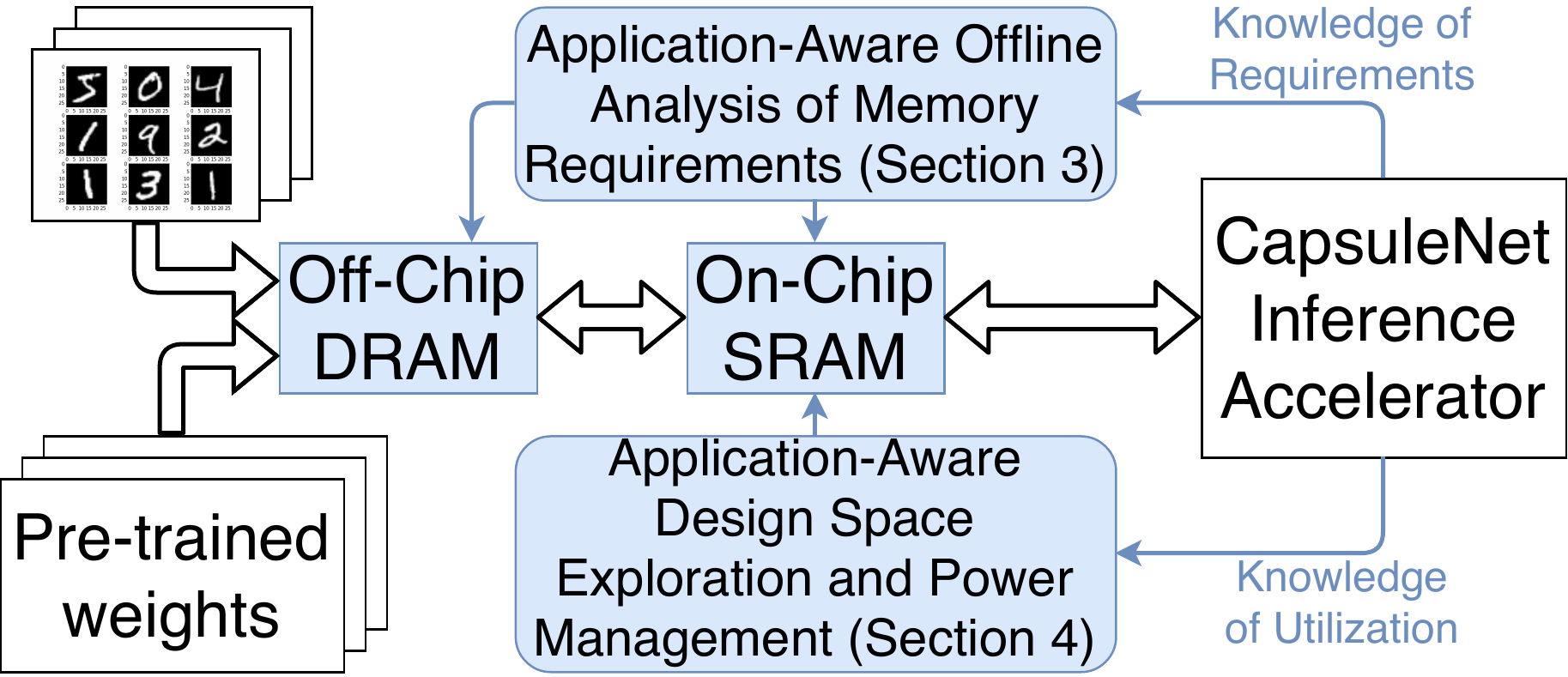}
	\vspace*{0mm}
	\caption{Overview of our CapStore Memory Design.}
	\label{fig:overview}
	\vspace*{0mm}
\end{figure}

\textbf{Research Challenges:} Traditional memory hierarchies of DNN accelerators are composed by an off-chip DRAM and an on-chip SRAM. An efficient design of a memory hierarchy requires exploration of several design parameters (like size, banks, partitions, etc.) for multiple levels, affecting each other, which makes it a very challenging problem. Intensive off-chip memory accesses reduce the energy efficiency and performance due to high access latency, while a large on-chip size may have a significant impact on the area and the leakage power. On the other hand, a large on-chip memory enhances high throughput computations, and a limited on-chip memory size is required to efficiently reduce the energy consumption. Further improvements can be achieved by systematic partitioning and power-gating (i.e., using sleep transistors \cite{ref:powergating} to switch-off the power supply of the correspondent memory sectors) under different memory usage scenarios. Note, the power-gating technique comes with the cost of wakeup energy and latency overhead. However, their usage is beneficial to significantly reduce the leakage power. To overcome the above challenges, \textit{a key information to consider could be the application-specific knowledge (i.e., different processing aspects of CapsuleNet inference)}. Such an application-aware design of memory hierarchy and the power management of the on-chip memory may bear the potential to provide significant energy savings compared to the traditional memory designs, while keeping high throughput, as we will demonstrate in this paper. The overview of our work is depicted in \Cref{fig:overview}, where the blue-colored boxes represent our novel contributions, as discussed below.

\vspace*{1mm}
\noindent
\textbf{Our Novel Contributions:}


\begin{enumerate}
    \item \textbf{Memory Analysis} (\Cref{sec:analysis}): we perform an extensive analysis of the memory requirements (size, accesses), performance and energy, for every operation of the CapsuleNet inference.
    \item \textbf{On-Chip Memory Architecture for CapsuleNet Accelerators} (\Cref{subsec:mem_model}): we propose CapStore, a multi-banked on-chip memory that can be partitioned into multiple sectors to support sector-level power-gating.
    \item \textbf{Design Space Exploration} (\Cref{subsec:design_space_exp}): we explore the key parameters of the memory architecture that leverage tradeoffs between memory partitionings, area and energy consumption, exploiting application-specific data.
    \item \textbf{Application-Aware Power Management} (\Cref{subsec:power_management}): it exploits the processing flow of the CapsuleNet inference and the architectural parameters of the accelerator and memory connections, to devise a sector-level power-gating, to further reduce the static power.
    \item \textbf{Implementation and Evaluation} (\Cref{sec:results}): we implement the CapsuleNet accelerator in a 32nm CMOS technology. 
    We compare the results in terms of area and energy consumption for different on-chip memory architectures, and benchmark against state-of-the-art design.
\end{enumerate}

Before proceeding to the technical sections, we present an overview of the CapsuleNets in \Cref{sec:background}, to a level of details which is necessary to understand the contributions of this paper.

\vspace*{-2mm}
\section{Background: CapsuleNets}
\label{sec:background}

The work by Hinton et al. \cite{ref:trans_autoencoder} showed the potential of CapsuleNets, particular types of DNNs where the basic units across the layers are capsules, i.e., multi-dimensional elements. The differences between CapsuleNets and DNNs are described in \Cref{subsec:capsnet_DNN}. Following the work of \cite{ref:dyn_routing}, which showed the efficient routing-by-agreement algorithm, CapsuleNets had become even more attractive for the communitity, thus demanding an effort from the hardware side to support such required processing. For this purpose, recently an accelerator for CapsuleNet inference was proposed in the work of \cite{ref:capsacc_arxiv}, whose architecture is described in \Cref{subsec:capsnet_accel}.

\vspace*{-2mm}
\subsection{Inference on CapsuleNets vs. DNNs}
\label{subsec:capsnet_DNN}

\begin{figure}[t]
	\centering
	\includegraphics[width=\linewidth]{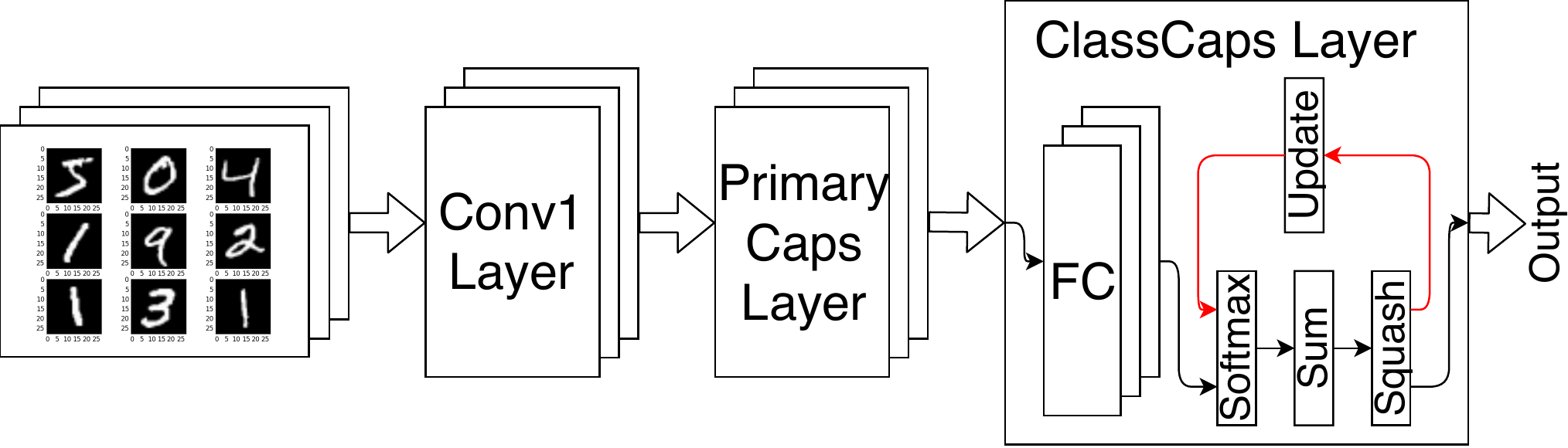}
	\vspace*{0mm}
	\caption{Architecture of the CapsuleNet for Inference \cite{ref:dyn_routing}.}
	\label{fig:capsnet_feedback}
	\vspace*{0mm}
\end{figure}

As compared to traditional DNNs, a CapsuleNet has: 
\begin{itemize}
    \item \textbf{Capsule:} a multi-dimensional neuron, which is able to encapsulate hierarchical information of multiple features (position, scale, orientation, etc.).
    \item \textbf{Squash activation function:} a multi-dimentional non-linear function, which efficiently fit for the prediction vector.
    \item \textbf{Routing-by-agreement:} an algorithm to learn the connection between two subsequent Capsule layers. \textit{It is an iterative algorithm, which iterates over a defined number of routing iterations}.
\end{itemize}

The last point is a very challenging aspect from the hardware perspective, as it means that \textit{there is a feedback loop in the inference path} (highlighted by the red-colored arrows in \Cref{fig:capsnet_feedback}). This property implies that it is more difficult to massively parallelize and pipeline the accelerator to compute such operations.

\vspace*{-2mm}
\subsection{CapsuleNet Accelerator}
\label{subsec:capsnet_accel}

\begin{figure}[t]
\centering
\vspace*{0mm}
\subfloat[]{
\includegraphics[width=.9\linewidth]{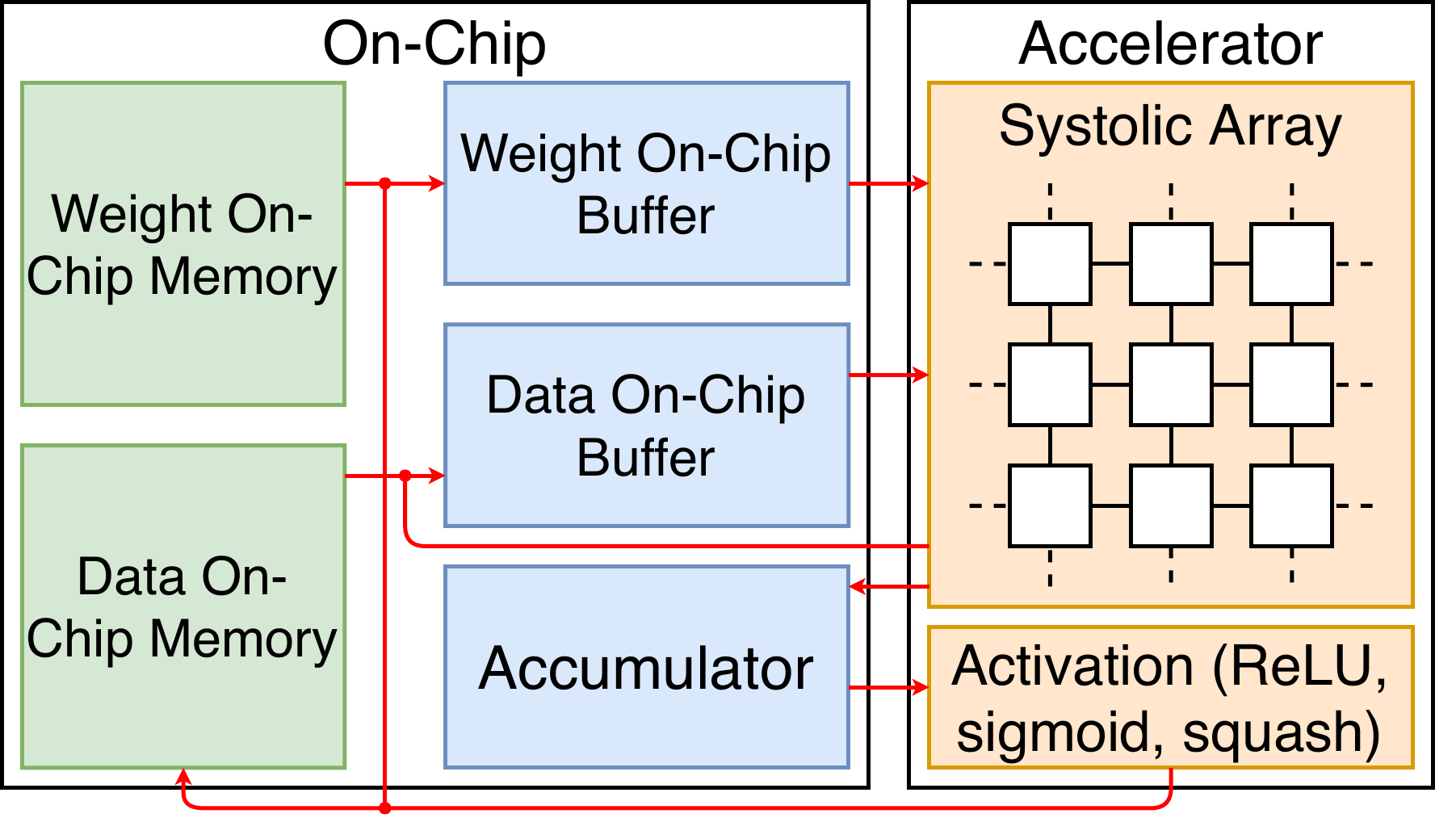}
\vspace*{-7mm}
\label{fig:capsacc}}
\vspace*{0mm}
\subfloat[]{
\includegraphics[width=.9\linewidth]{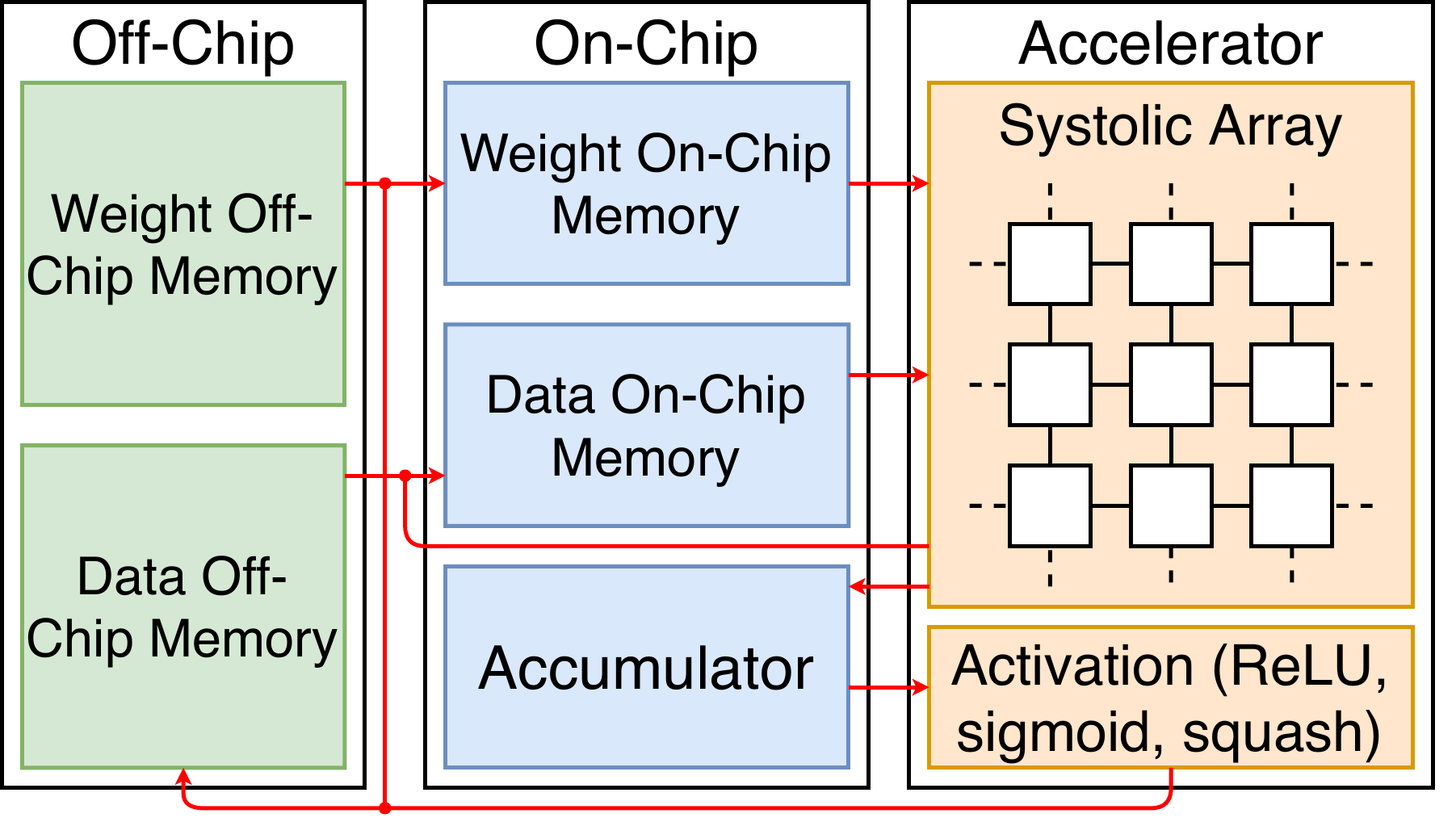}
\vspace*{-7mm}
\label{fig:capsacc_dac}}
\vspace*{-3mm}
\caption{Architectural view of the CapsuleNet inference accelerator. (a) Architecture proposed in \cite{ref:capsacc_arxiv}. (b) Our proposed architecture.}
\label{fig:capsacc_architectures}
\vspace*{0mm}
\end{figure}

\Cref{fig:capsacc} shows the architecture of the specialized accelerator for CapsuleNet inference \cite{ref:capsacc_arxiv}. The core of the processing unit is the systolic array (16x16 Processing Elements), for efficiently parallelizing the computation. Systolic array-based architectures have been also used for specialized DNN accelerators \cite{ref:Eyeriss} \cite{ref:TPU}. The accumulator stores the partial sums and computes further internal additions when required. The activation unit computes the activation functions needed. The dedicated connections between accelerator and memories allow data and weight reuse, properties which are particularly effective for the routing-by-agreement computation.

In the original paper \cite{ref:capsacc_arxiv}, all the memory elements are on-chip, forming an overall size of 8MB. Since for memory constrained systems such size can potentially exceed the limits, we decided to break the memory hierarchy into an on-chip SRAM, followed by an off-chip DRAM, as shown in the blue-colored boxes and green-colored boxes of \Cref{fig:capsacc_dac}, respectively. This solution can potentially generalize the problem for different applications and more complex CapsuleNet architectures. For this purpose, we adopt the following policies to define the sizes and the communications between off-chip and on-chip memories:
\begin{itemize}
    \item Minimize the off-chip memory accesses.
    \item Keep the same latency and throughput, as compared to having all the memory on-chip.
    \item Minimize the on-chip memory size.
\end{itemize}
The aforementioned constraints and limitations motivate our analysis of the resource requirements, which is presented in \Cref{sec:analysis}.

\vspace*{-2mm}
\section{Analysis: CapsuleNet Resource Requirements}
\label{sec:analysis}

We perform an extensive analysis to identify the resource requirements for computing the inference on CapsuleNets. As a case study, we investigate the CapsuleNet architecture described by \cite{ref:dyn_routing}, which performs MNIST \cite{ref:MNIST} classification. First, in \Cref{subsec:perf_mem_breakdown} we analyze the performance and the on-chip memory requirements for every operation of the CapsuleNet inference. Then, we analyze the on-chip read and write accesses for every operation of the inference. Finally, in \Cref{subsec:energy_breakdown} we compute the energy breakdown, between the accelerator and the components of the memory. 

\vspace*{-2mm}
\subsection{Performance, Memory Usage and Accesses}
\label{subsec:perf_mem_breakdown}

\begin{figure}[t]
    \centering
    \begin{figure}[H]
    \vspace*{-3mm}
    \includegraphics[width=\linewidth]{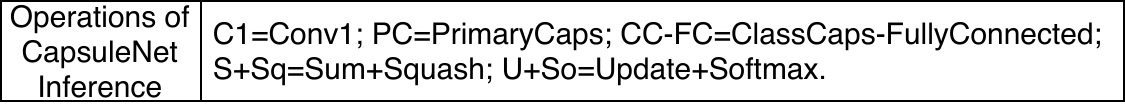}
    \vspace*{-8mm}
    \end{figure}
	\subfloat[]{
	\begin{minipage}[t]{.08\linewidth}
	\hfill
	\end{minipage}
	\begin{minipage}[t]{.92\linewidth}
	\includegraphics[width=\linewidth]{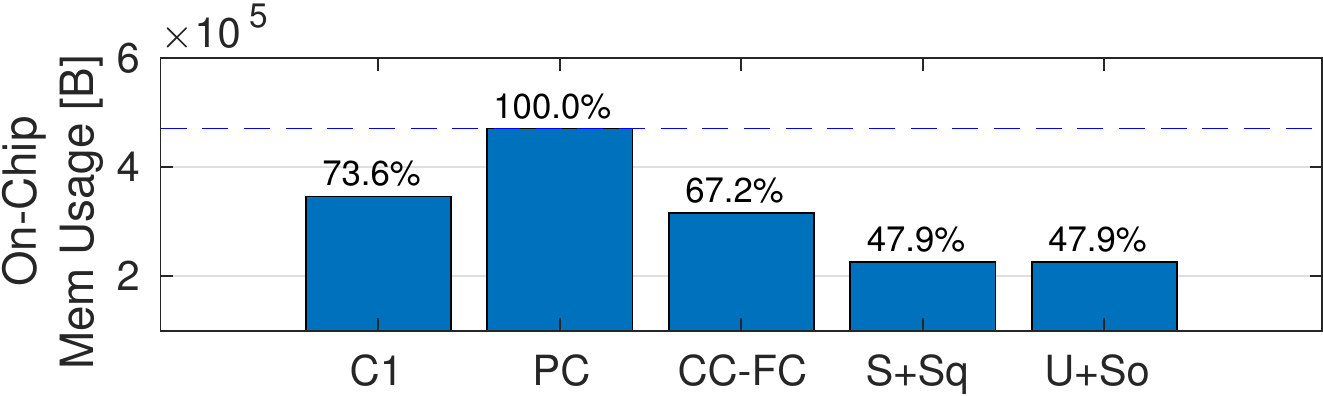}
	\vspace*{-10mm}
	\label{fig:mem_analysis}
	\end{minipage}} \\
	\vspace*{-2mm}
	\subfloat[]{
	\begin{minipage}[t]{.04\linewidth}
	\hfill
	\end{minipage}
	\begin{minipage}[t]{.96\linewidth}
	\includegraphics[width=\linewidth]{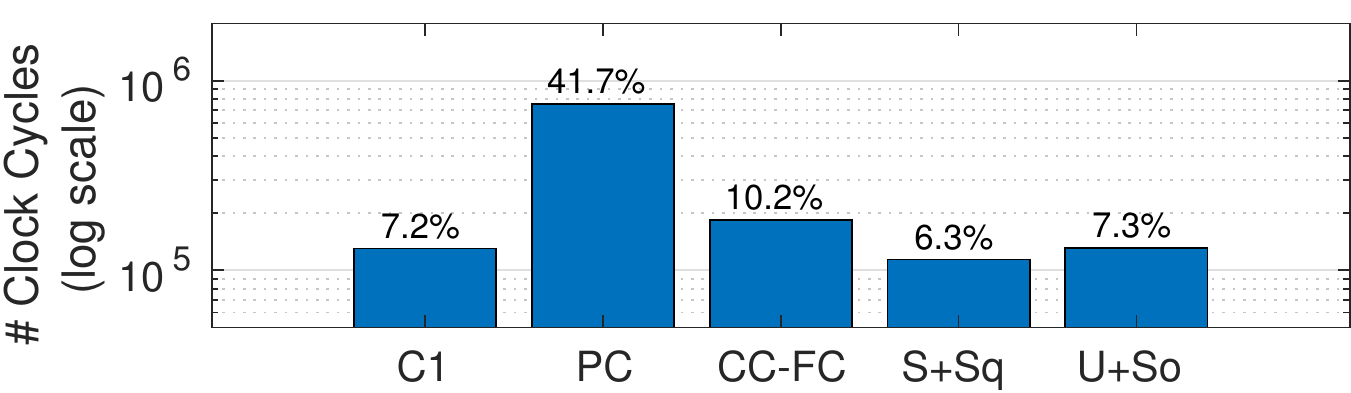}
	\vspace*{-6mm}
	\label{fig:cyc_analysis}
	\end{minipage}} \\
	\vspace*{-2mm}
	\subfloat[]{
	\includegraphics[width=\linewidth]{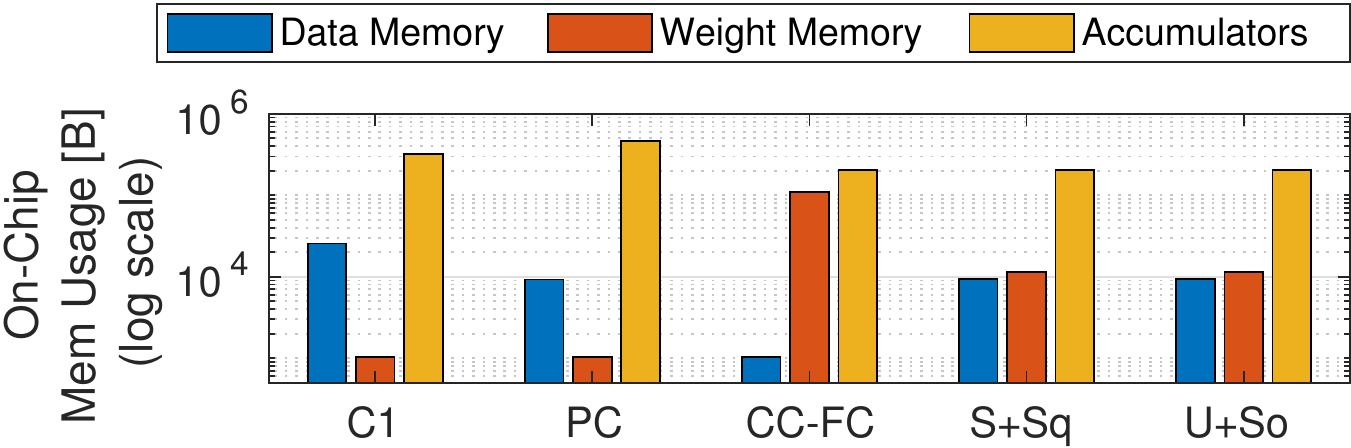}
	\label{fig:mem_separated}} \\
	\vspace*{-2mm}
	\subfloat[]{
	\begin{minipage}[t]{.04\linewidth}
	\hfill
	\end{minipage}
	\begin{minipage}[t]{.96\linewidth}
	\includegraphics[width=\linewidth]{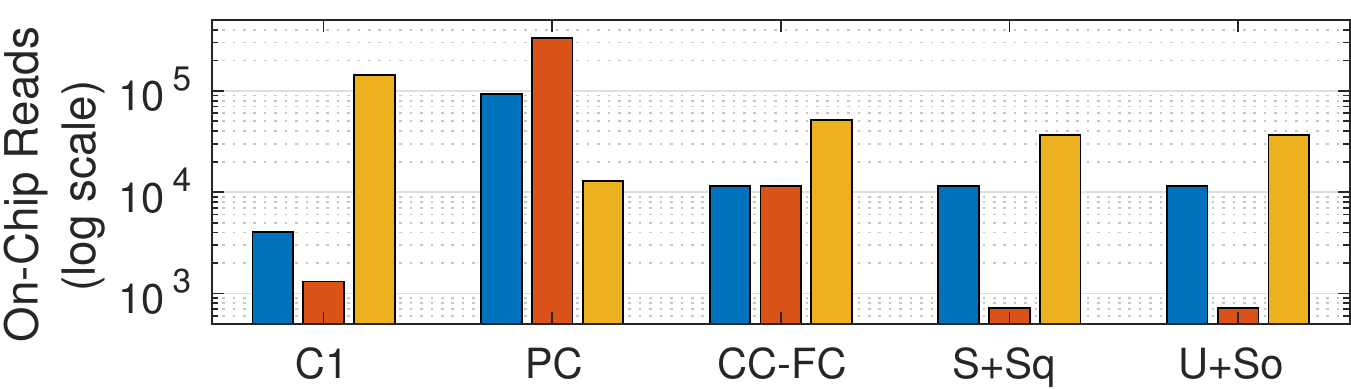}
	\vspace*{-6mm}
	\label{fig:mem_reads}
	\end{minipage}} \\
	\vspace*{-2mm}
	\subfloat[]{
	\begin{minipage}[t]{.04\linewidth}
	\hfill
	\end{minipage}
	\begin{minipage}[t]{.96\linewidth}
	\includegraphics[width=\linewidth]{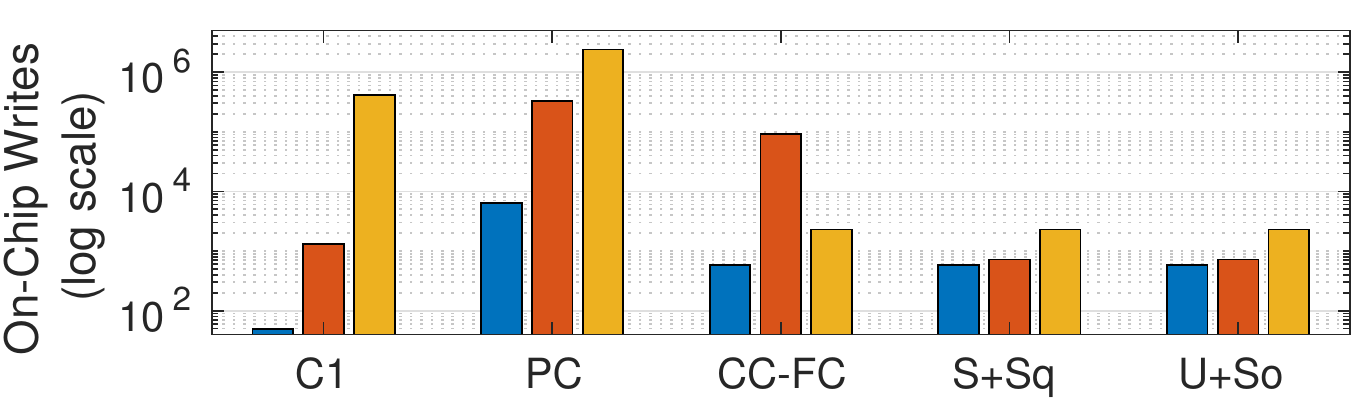}
	\vspace*{-6mm}
	\label{fig:mem_writes}
	\end{minipage}}
	\vspace*{-2mm}
	\caption{Resource requirements of each operation of the CapsuleNet inference: (a) On-chip memory requirements. (b) Number of clock cycles. (c) On-chip memory requirements of each memory component. (d) On-chip memory reads for each memory component. (e) On-chip memory reads for each memory component.}
	\label{fig:mem_perf}
	\vspace*{0mm}
\end{figure}

Considering the design policies discussed in \Cref{subsec:capsnet_accel}, we analyze the on-chip memory requirements for each operation of the CapsuleNet inference. The results are shown in \Cref{fig:mem_analysis}. The overall size is determined by the operation which requires the largest amount of memory (PrimaryCaps layer in our case). For this configuration, the on-chip memory is shared between data, weight and accumulator items. The dashed line represents the maximum value, and for each operation we display the percentage of utilization.

\Cref{fig:cyc_analysis} reports the number of clock cycles required to compute each operation of the CapsuleNet inference. Note, the last two operations (Sum+Squash and Update+Sum) are executed at each routing iteration (i.e., 3 times in our example). If we combine the results of \Cref{fig:mem_analysis,fig:cyc_analysis}, we notice that, potentially, a significant amount of leakage energy can be saved by power-gating part of the on-chip memory, when the utilization is below 100\%. This idea will be described and implemented in \Cref{subsec:power_management}.

\Cref{fig:mem_separated} presents the memory requirements for each memory component (data memory, weight memory and accumulators). Such analysis enables the design space exploration of an application-aware memory architecture (CapStore), which explores the possibility to handling separate the memory components of the on-chip memory. It will be discussed in \Cref{subsec:design_space_exp}. The accumulator size is higher than data and weight memory for each operation, because it must store the temporary partial sums of different output feature maps. Data and weight memory requirements, however, vary significantly across different operations. In the first two layers, the weight memory requirements are quite low as compaerd to the other stages, because the architecture can efficiently employ weight reuse in convolutional networks. In the ClassCaps layer, however, the data memory is low, because data reuse is efficient. Weight reuse is also more efficient in the last two operations, as compared to the third one.

\Cref{fig:mem_reads,fig:mem_writes} show the read and write accesses, respectively, for each operation $i$ of the inference, i.e., C1, PC and CC-FC. These values are needed to compute the energy consumptions of the memories in the following sections. Note, the off-chip accesses are not reported in the graphs for space reasons. Their values can be easily computed using the \Cref{eq:offchip_read,eq:offchip_write}, which are valid for the first three operations. In the last two operations, the off-chip memory is not accessed: all the values that have to be saved during the routing-by-agreement are stored on-chip.

\vspace*{-4mm}
\begin{equation}
    {\scriptstyle \left( \# Reads_{off-chip} \right) _{i} = \left( \# Writes_{weight-mem} + \# Writes_{data-mem} \right) _{i} }
    \label{eq:offchip_read}
\end{equation}
\begin{equation}
    {\scriptstyle \left( \# Writes_{off-chip} \right) _{i} = \left( \# Reads_{data-mem} \right) _{i+1} }
    \label{eq:offchip_write}
\end{equation}

\subsection{Energy Breakdown}
\label{subsec:energy_breakdown}

\begin{figure}[t]
\centering
\begin{minipage}[t]{.47\linewidth}
\vspace*{0mm}
\subfloat[]{
\includegraphics[width=\linewidth]{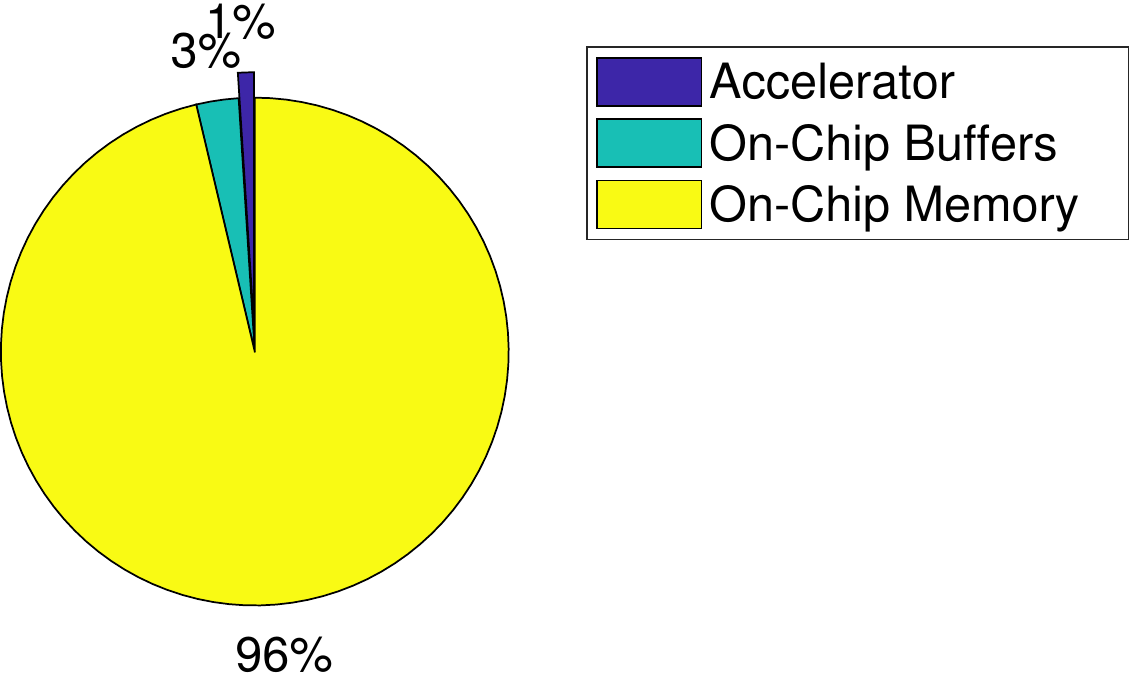}
\label{fig:energy_breakdown_sram}}
\end{minipage}
\hfill
\begin{minipage}[t]{.52\linewidth}
\vspace*{0mm}
\subfloat[]{
\includegraphics[width=\linewidth]{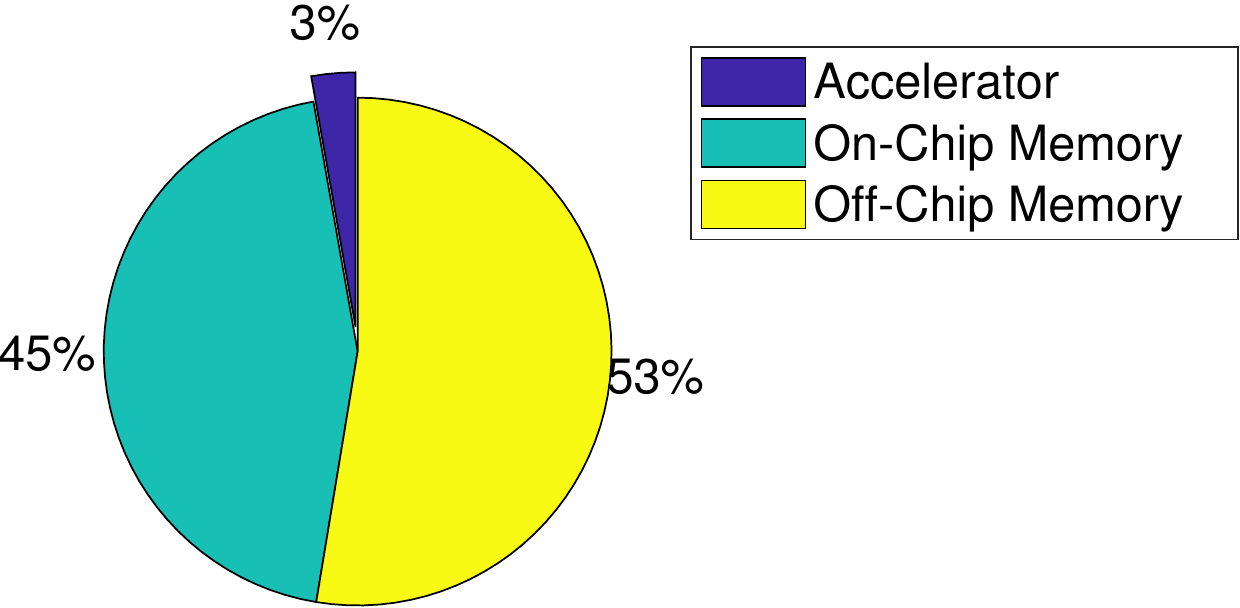}
\label{fig:energy_breakdown_dram}}
\end{minipage}
\vspace*{-4mm}
\caption{Energy breakdown of the different components of the CapsuleNet Inference Architecture: considering (a) all on-chip, as employed in \cite{ref:capsacc_arxiv} and (b) a memory hierarchy composed by on-chip and off-chip memories.}
\label{fig:energy_breakdown}
\vspace*{0mm}
\end{figure}

\begin{figure*}[t]
\begin{minipage}[t]{.42\linewidth}
\begin{figure}[H]
	\centering
	\includegraphics[width=\linewidth]{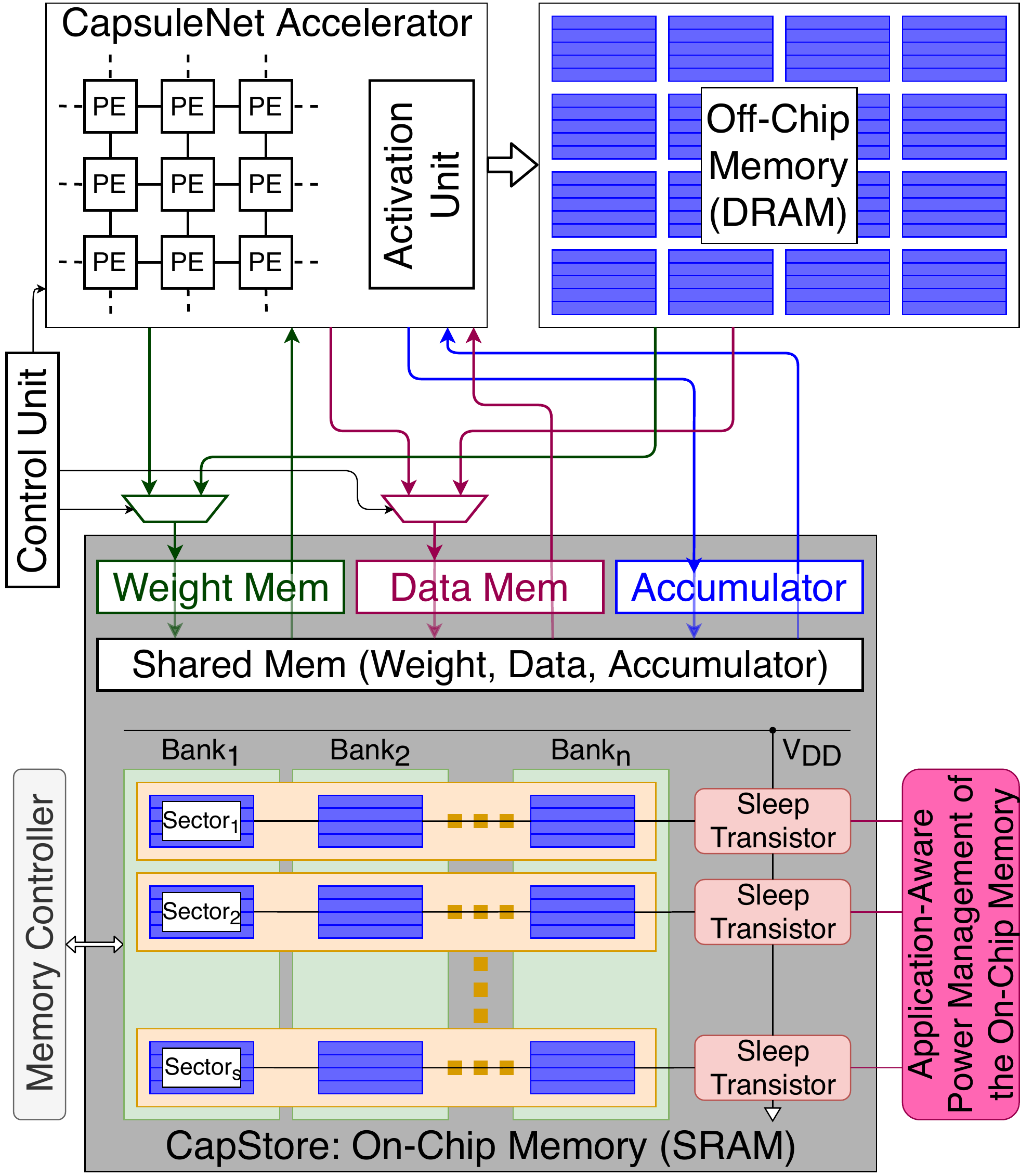}
	\vspace*{-5mm}
	\caption{Architectural model of the complete CapsuleNet architecture, with a focus on our CapStore on-chip memory.}
	\label{fig:memory_model}
	\vspace*{-2mm}
\end{figure}
\end{minipage}
\hfill
\begin{minipage}[t]{.57\linewidth}
\begin{figure}[H]
\centering
\vspace*{0mm}
\begin{minipage}[t]{.40\linewidth}
\subfloat[]{
\includegraphics[width=\linewidth]{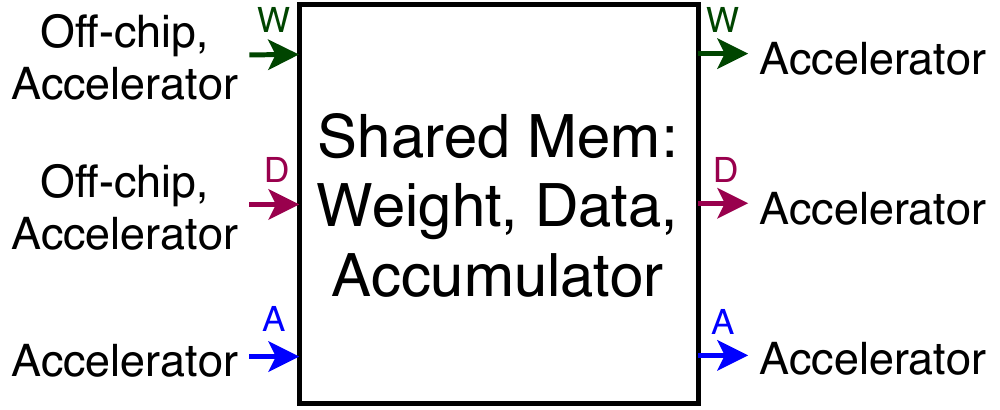}
\label{fig:multiport_mem}}
\vspace*{-2mm}
\subfloat[]{
\includegraphics[width=\linewidth]{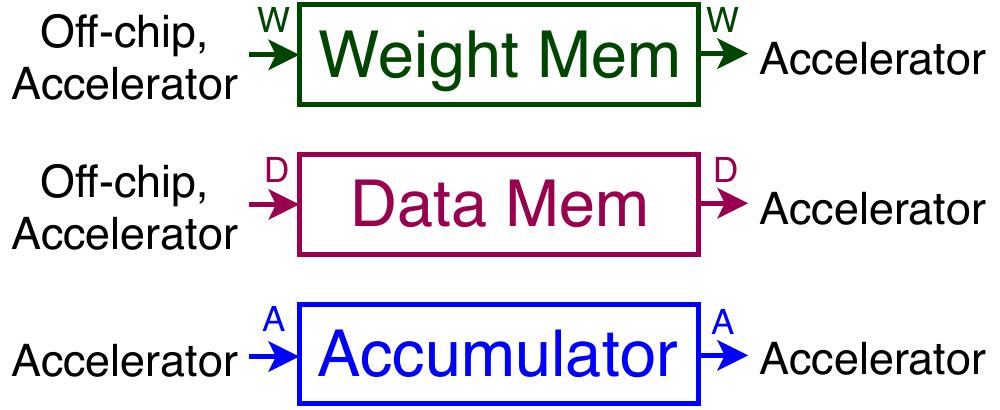}
\label{fig:separate_element}}
\end{minipage}
\hfill
\begin{minipage}[t]{.59\linewidth}
\vspace*{0mm}
\subfloat[]{
\includegraphics[width=\linewidth]{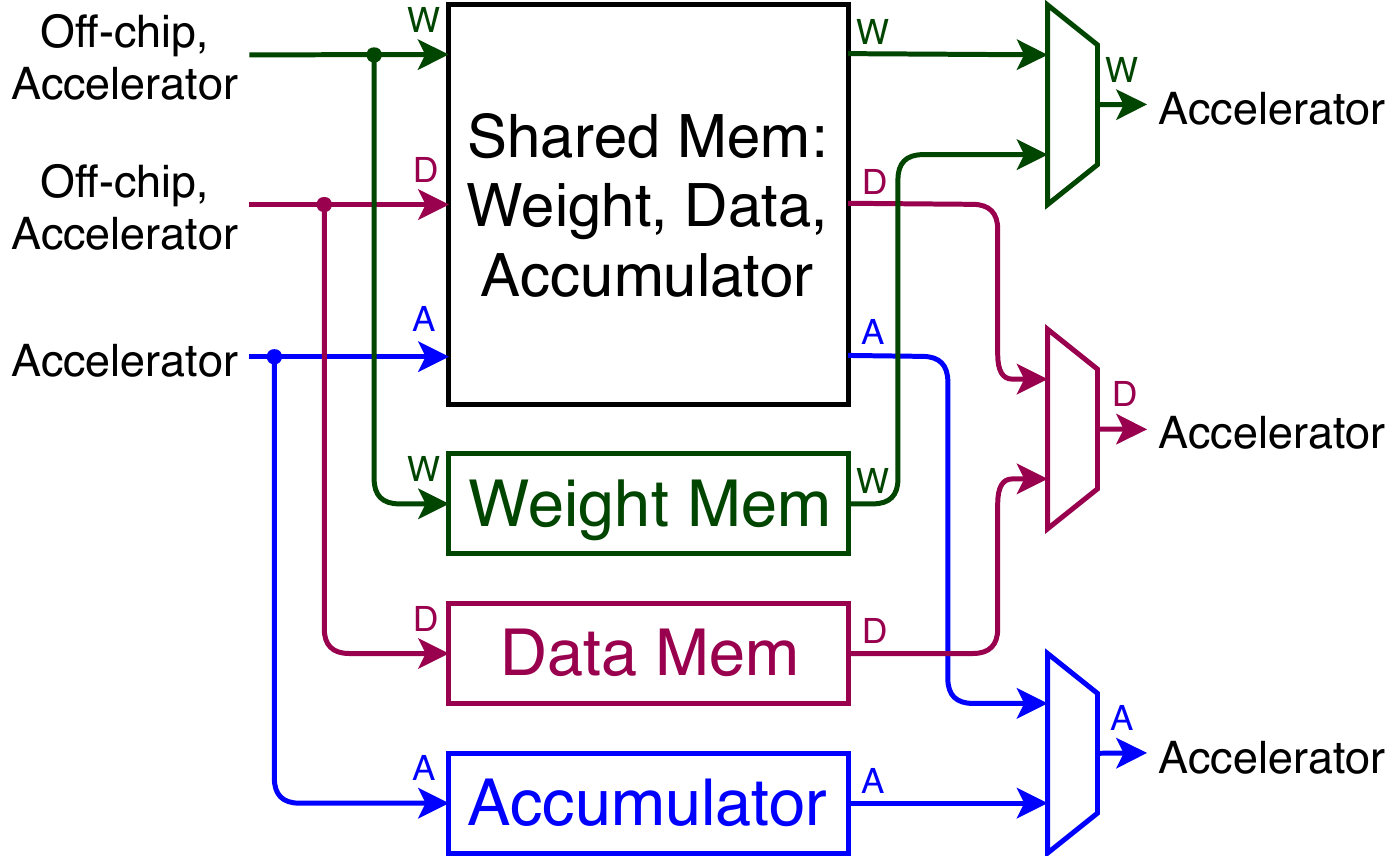}
\label{fig:hybrid_mem}}
\end{minipage}
\vspace*{-4mm}
\caption{Schematic of Different Architectures of the On-Chip Memory of the CapsuleNet Accelerator that are evaluated in our Application-Aware Design Space Exploration: (a) Shared Multi-Port Memory. (b) Separated Memory. (c) Hybrid Memory (Shared and Separated).}
\label{fig:memory_architectures}
\vspace*{-4mm}
\end{figure}
\begin{table}[H]
\resizebox{\linewidth}{!}{%
\begin{tabular}{c|c|c|c|c|c|c|c|c|c|c|c|c|}
\cline{2-13}
\textbf{} & \multicolumn{3}{c|}{\textbf{Shared Mem}} & \multicolumn{3}{c|}{\textbf{Weight Mem}} & \multicolumn{3}{c|}{\textbf{Data Mem}} & \multicolumn{3}{c|}{\textbf{Accumulator Mem}} \\ 
\textbf{} & \multicolumn{1}{|c}{\textbf{Size (B)}} & \multicolumn{1}{c}{\textbf{N}} & \multicolumn{1}{c|}{\textbf{S}} & \multicolumn{1}{|c}{\textbf{Size (B)}} & \multicolumn{1}{c}{\textbf{N}} & \multicolumn{1}{c|}{\textbf{S}} & \multicolumn{1}{|c}{\textbf{Size (B)}} & \multicolumn{1}{c}{\textbf{N}} & \multicolumn{1}{c|}{\textbf{S}} & \multicolumn{1}{|c}{\textbf{Size (B)}} & \multicolumn{1}{c}{\textbf{N}} & \multicolumn{1}{c|}{\textbf{S}} \\ \hline
\multicolumn{1}{|c|}{SMP} & 471040 & 16 & 1 & - & - & - & - & - & - & - & - & - \\ \hline
\multicolumn{1}{|c|}{\color{red}PG\color{black}-SMP} & 471040 & 16 & \color{red}256\color{black} & - & - & - & - & - & - & - & - & - \\ \hline
\multicolumn{1}{|c|}{SEP} & - & - & - & 110592 & 16 & 1 & 25600 & 16 & 1 & 460800 & 16 & 1 \\ \hline
\multicolumn{1}{|c|}{\color{red}PG\color{black}-SEP} & - & - & - & 110592 & 16 & \color{red}64\color{black} & 25600 & 16 & \color{red}16\color{black} & 460800 & 16 & \color{red}128\color{black} \\ \hline
\multicolumn{1}{|c|}{HY} & 264192 & 16 & 1 & 1024 & 16 & 1 & 1024 & 16 & 1 & 204800 & 16 & 1 \\ \hline
\multicolumn{1}{|c|}{\color{red}PG\color{black}-HY} & 264192 & 16 & \color{red}128\color{black} & 1024 & 16 & 1 & 1024 & 16 & 1 & 204800 & 16 & 1 \\ \hline
\end{tabular}%
}
\caption{Sizes, banks and sectors for each CapStore memory architecture organization. The prefix `PG-' stands for power-gating.}
\label{tab:sizes}
\vspace*{-6mm}
\end{table}
\end{minipage}
\vspace*{0mm}
\end{figure*}

In this stage, we compute the energy consumption of the complete architecture. We develop two different versions:

\begin{enumerate}[label=(\alph*)]
    \item Fig. \ref{fig:capsacc}: accelerator (composed by systolic array, activation unit and control unit), on-chip buffers (data buffer, weight buffer and accumulator) and on-chip memory (data memory and weight memory). This is the architecture proposed in \cite{ref:capsacc_arxiv}, where the on-chip memory has size equal to 8MB.
    \item Fig. \ref{fig:capsacc_dac}: accelerator, on-chip and off-chip memories. The sizes are derived from the requirements analyzed in \Cref{subsec:perf_mem_breakdown}.
\end{enumerate}

The energy breakdowns are shown in \Cref{fig:energy_breakdown}. Note, the results relative to the accelerator and the on-chip buffers have been obtained by synthesizing the CapsuleNet accelerator of \cite{ref:capsacc_arxiv} in a 32nm CMOS technology, while on-chip and off-chip memory values have been extracted using CACTI-P \cite{ref:CACTI-P}.

This analysis shows that, by organizing the memory in a different hierarchy, we can already save 66\% of the total energy, as compared to the state-of-the-art architecture \cite{ref:capsacc_arxiv}. Moreover, since the on-chip memory consumes 45\% of the total energy, an application-aware power management (discussed in \Cref{subsec:power_management}) can potentially have a significant impact on the overall energy consumption.

\vspace*{-2mm}
\subsection{Key Observations from our Analyses}
\label{subsec:key_observations}

From the analyses performed in \Cref{subsec:perf_mem_breakdown,subsec:energy_breakdown}, we derive the following key observations:

\begin{itemize}
\item Most of the energy is consumed by the (on-chip and off-chip) memory, as compared to the computational array of the accelerator.
\item An application-aware memory hierarchy, composed by an on-chip SRAM and an off-chip DRAM, can save up to 66\% of energy, without compromising the throughput, as compared to having a fully on-chip memory organization.
\item The utilization of the on-chip memory is variable, depending upon the operation of the CapsuleNet inference. Thus, applying power-gating to the non utilized sectors can potentially further reduce the energy consumption.
\item Partitioning the on-chip memory into separated chips (for data, weight and accumulator) can be beneficial for storing velues and feeding the accelerator in an efficient way.
\end{itemize}

\vspace*{-3mm}
\section{CapStore: On-Chip Memory Design and Management}
\label{sec:design}

\vspace*{-1mm}
\subsection{Memory Models}
\label{subsec:mem_model}

In this section, we present the memory models used in our work. Our reference design of the CapStore on-chip memory is shown in \Cref{fig:memory_model}. The on-chip memory is connected to the CapsuleNet accelerator and to the off-chip memory through dedicated bus lines. We design our CapStore memory as partitioned into N banks, and each bank into S equally-sized sectors. All the sectors with the same index, across different banks, are connected through a power-gating circuitry to the same sleep transistor. This implies that each sleep transistor is responsible for power-gating N sectors, one for each bank. The sleep transistors are connected to our application-aware power management unit, which gives the control signals to handle \textit{ON} $\leftrightarrow$ \textit{OFF} transitions. Since we perform power-gating on the on-chip SRAM and we do not need data retention, our model consists of just two sleep modes, i.e., \textit{ON} (full swing voltage) and \textit{OFF} (zero voltage). All the intermediate sleep modes (data retentive with reduced voltage) has not been considered in our model, since they are not useful for our scenarios. The transitions between sleep modes come at a cost of a certain wakeup energy and latency overhead. However, the usage of power-gating leverages the tradeoff between wakeup overhead and static (leakage) power savings. Note, our memory model can be generalized for each memory size and parallelism, including multi-port memories.

We design 3 different on-chip memory architectures:
\begin{enumerate}[label=(\alph*)]
    \item \textbf{Fig. \ref{fig:multiport_mem} - Shared Multi-Port Memory (SMP):} the on-chip memory is shared across a multi-port memory, which has 3 ports (for weight memory, data memory and accumulator).
    \item \textbf{Fig. \ref{fig:separate_element} - Separated Memory (SEP):} weight memory, data memory and accumulator are separate memories.
    \item \textbf{Fig. \ref{fig:hybrid_mem} - Hybrid Memory (HY):} a combination of a shared multi-port and three separated memories.
\end{enumerate}

\vspace*{-3mm}
\subsection{Application-Aware Design Space Exploration}
\label{subsec:design_space_exp}

Having defined the memory models, we now design the number of banks (N), the number of sectors (S) per bank, the organization, the parallelism and the power-gating management. In this section we cover all the aforementioned aspects, except for the power-gating, which will be discussed in \Cref{subsec:power_management}.

The knowledge of the application plays a key role for designing the on-chip memory. We explore different configurations of the memory architecture and, for each case, we compute area and energy consumption (see \Cref{sec:results}). Different levels of abstraction of application-aware knowledge are employed. The derivations are explaied in the following considerations.

\begin{itemize}
    \item \textbf{Architecture:} the parallelism of the systolic array of the CapsuleNet architecture (16x16 processing elements) suggests to employ 16 banks.
    \item \textbf{Overall utilization:} the memory requirements (worst case of \Cref{fig:mem_analysis}) suggest the size of the SMP memory.
    \item \textbf{Utilization for different elements:} the worst case memory requirements of \Cref{fig:mem_separated} and the organization into different blocks (data memory, weight memory and accumulator) suggest the sizes of the SEP memory. The minimum utilization of the memory in \Cref{fig:mem_separated} suggests the sizes of the separated memories in the HY architecture. The size of the shared multi-port memory of the HY architecture is suggested by the difference between the worst case utilization and the sum of the sizes of the separated memories.
    \item \textbf{Utilization across different layers:} \Cref{fig:mem_analysis,fig:mem_separated} suggest the sector size, in order to apply power-gating to the unused sectors of the memory.
\end{itemize}

We design six different versions of the CapStore on-chip memory, one for each memory architecture presented in \Cref{fig:memory_architectures}, with or without enabling power-gating (denoted by the prefix -PG). The different architecture organizations are reported in \Cref{tab:sizes}. Note, when power-gating is not enabled, the memory architectures are slightly different: the hardware overhead composed by the sleep transistors and the power management unit (PMU) is missing. Therefore, for such architectures, each bank contains only 1 sector.

\vspace*{-2mm}
\subsection{Application-Aware Power Management}
\label{subsec:power_management}

\begin{figure}[t]
	\centering
	\includegraphics[width=.7\linewidth]{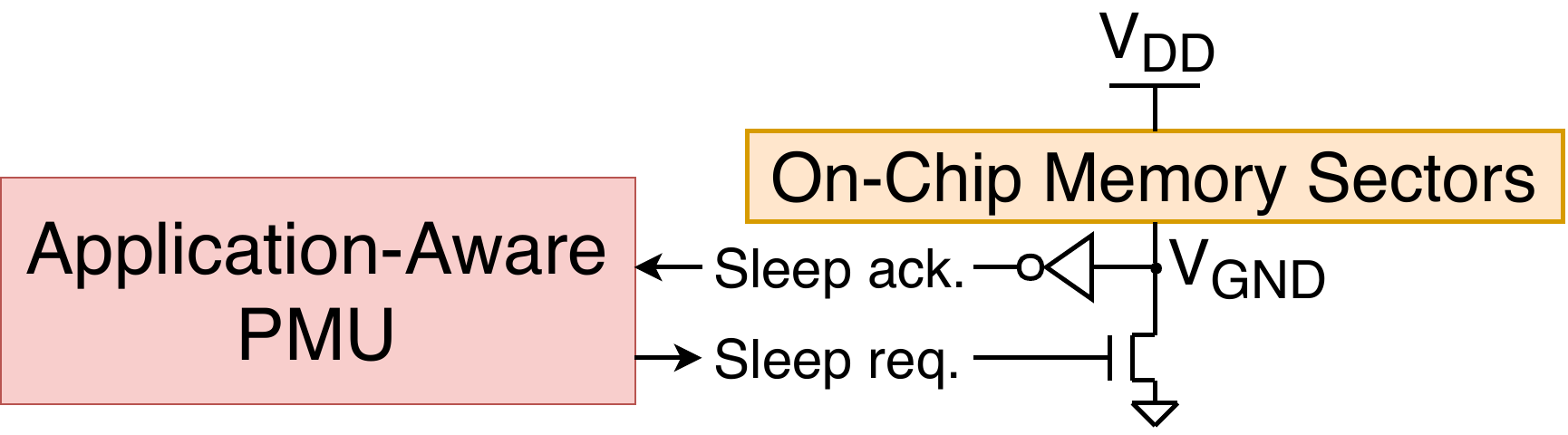}
	\vspace*{0mm}
	\caption{Circuit-level schematic of the power-gating circuit, using a footer sleep transistor connected to the PMU.}
	\label{fig:sleep_circuit}
	\vspace*{0mm}
\end{figure}

\begin{figure}[t]
	\centering
	\includegraphics[width=\linewidth]{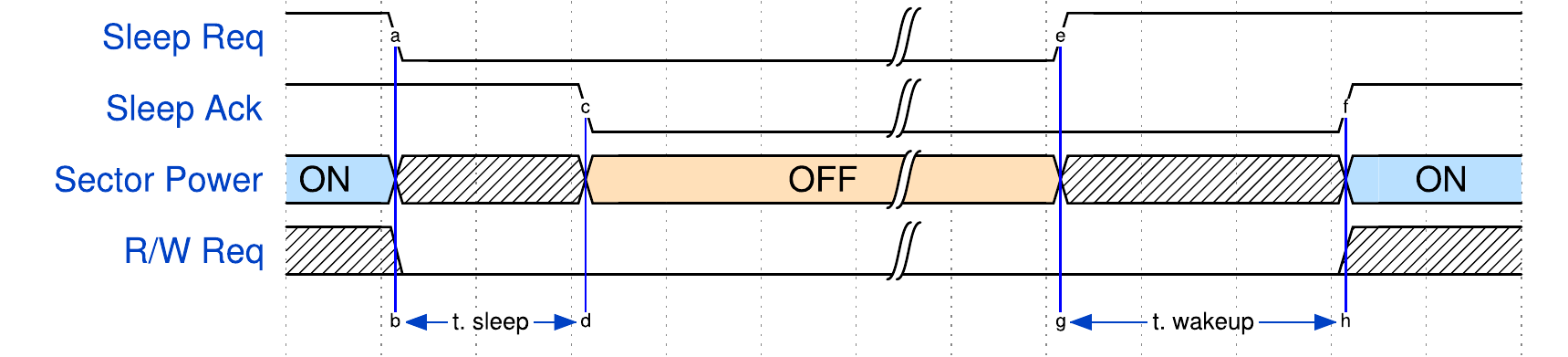}
	\vspace*{-6mm}
	\caption{Timing diagram of a complete sleep cycle of a sector.}
	\label{fig:sleep_cycle}
	\vspace*{0mm}
\end{figure}

Our application-aware PMU is responsible to give the control signals to the sleep transistors, according to the utilization of the memory, derived from the \Cref{fig:mem_analysis,fig:mem_separated}. The additional hardware circuitry, apart from the PMU itself, is represented by the sleep transistors. A simple schematic showing how one sleep transistor is connected to its relative sectors of the memory is depicted in \Cref{fig:sleep_circuit}. The sleep request is followed by the acknowledge signal, forming a 2-way handshake protocol. The timing diagram of a complete sleep cycle (\textit{ON} $\rightarrow$ \textit{OFF} $\rightarrow$ \textit{ON}) is shown in \Cref{fig:sleep_cycle}.

\vspace*{-2mm}
\section{CapStore Architecture Evaluation}
\label{sec:results}

\vspace*{-1mm}
\subsection{Area and Energy Consumption of the On-Chip Memory}
\label{subsec:results_onchip}

\begin{table*}[t]
\resizebox{.85\linewidth}{!}{%
\begin{tabular}{c|c|c|c|c|c|c|c|c|}
\cline{2-9}
\textbf{} & \multicolumn{2}{c|}{\textbf{Shared Mem}} & \multicolumn{2}{c|}{\textbf{Weight Mem}} & \multicolumn{2}{c|}{\textbf{Data Mem}} & \multicolumn{2}{c|}{\textbf{Accumulator Mem}} \\
\textbf{} & \multicolumn{1}{|c}{\textbf{Area {[}mm2{]}}} & \multicolumn{1}{c|}{\textbf{Energy {[}mJ{]}}} & \multicolumn{1}{|c}{\textbf{Area {[}mm2{]}}} & \multicolumn{1}{c|}{\textbf{Energy {[}mJ{]}}} & \multicolumn{1}{|c}{\textbf{Area {[}mm2{]}}} & \multicolumn{1}{c|}{\textbf{Energy {[}mJ{]}}} & \multicolumn{1}{|c}{\textbf{Area {[}mm2{]}}} & \multicolumn{1}{c|}{\textbf{Energy {[}mJ{]}}} \\ \hline
\multicolumn{1}{|c|}{All On-Chip {[}11{]}} & \multicolumn{1}{c|}{18.486} & \multicolumn{1}{c|}{38.6733} & \multicolumn{1}{c|}{} & \multicolumn{1}{c|}{} & \multicolumn{1}{c|}{} & \multicolumn{1}{c|}{} & \multicolumn{1}{c|}{} & \multicolumn{1}{c|}{} \\ \hline
\multicolumn{1}{|c|}{SMP} & \multicolumn{1}{c|}{11.4232} & \multicolumn{1}{c|}{8.7088} & \multicolumn{1}{c|}{-} & \multicolumn{1}{c|}{-} & \multicolumn{1}{c|}{-} & \multicolumn{1}{c|}{-} & \multicolumn{1}{c|}{-} & \multicolumn{1}{c|}{-} \\ \hline
\multicolumn{1}{|c|}{PG-SMP} & \multicolumn{1}{c|}{34.4412} & \multicolumn{1}{c|}{7.9194} & \multicolumn{1}{c|}{-} & \multicolumn{1}{c|}{-} & \multicolumn{1}{c|}{-} & \multicolumn{1}{c|}{-} & \multicolumn{1}{c|}{-} & \multicolumn{1}{c|}{-} \\ \hline
\multicolumn{1}{|c|}{SEP} & \multicolumn{1}{c|}{-} & \multicolumn{1}{c|}{-} & \multicolumn{1}{c|}{0.108034} & \multicolumn{1}{c|}{0.1659} & \multicolumn{1}{c|}{0.815363} & \multicolumn{1}{c|}{0.7136} & \multicolumn{1}{c|}{2.20981} & \multicolumn{1}{c|}{3.1603} \\ \hline
\multicolumn{1}{|c|}{\textbf{PG-SEP}} & \multicolumn{1}{c|}{-} & \multicolumn{1}{c|}{-} & \multicolumn{1}{c|}{0.514265} & \multicolumn{1}{c|}{\textbf{0.0447}} & \multicolumn{1}{c|}{1.64803} & \multicolumn{1}{c|}{\textbf{0.1364}} & \multicolumn{1}{c|}{3.9458} & \multicolumn{1}{c|}{\textbf{1.0109}} \\ \hline
\multicolumn{1}{|c|}{HY} & \multicolumn{1}{c|}{7.11157} & \multicolumn{1}{c|}{5.4014} & \multicolumn{1}{c|}{0.0215973} & \multicolumn{1}{c|}{0.0123} & \multicolumn{1}{c|}{0.0215973} & \multicolumn{1}{c|}{0.0190} & \multicolumn{1}{c|}{1.17416} & \multicolumn{1}{c|}{1.5467} \\ \hline
\multicolumn{1}{|c|}{PG-HY} & \multicolumn{1}{c|}{19.427} & \multicolumn{1}{c|}{3.8613} & \multicolumn{1}{c|}{0.0215973} & \multicolumn{1}{c|}{0.0123} & \multicolumn{1}{c|}{0.0215973} & \multicolumn{1}{c|}{0.0190} & \multicolumn{1}{c|}{1.17416} & \multicolumn{1}{c|}{1.5467} \\ \hline
\end{tabular}%
}
\caption{Area and energy consumption for different CapStore on-chip memory architectures.}
\label{tab:results_final}
\vspace*{0mm}
\end{table*}

\begin{figure}[t]
\vspace*{0mm}
    \centering
    \subfloat[]{
	\includegraphics[width=\linewidth]{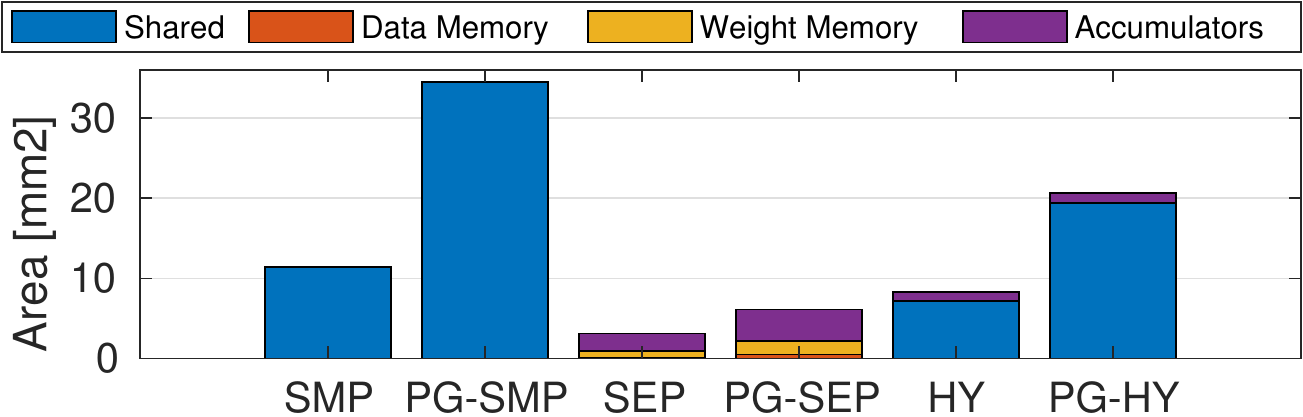}
	\label{fig:area_versions}} \\
	\vspace*{-2mm}
	\subfloat[]{
	\includegraphics[width=\linewidth]{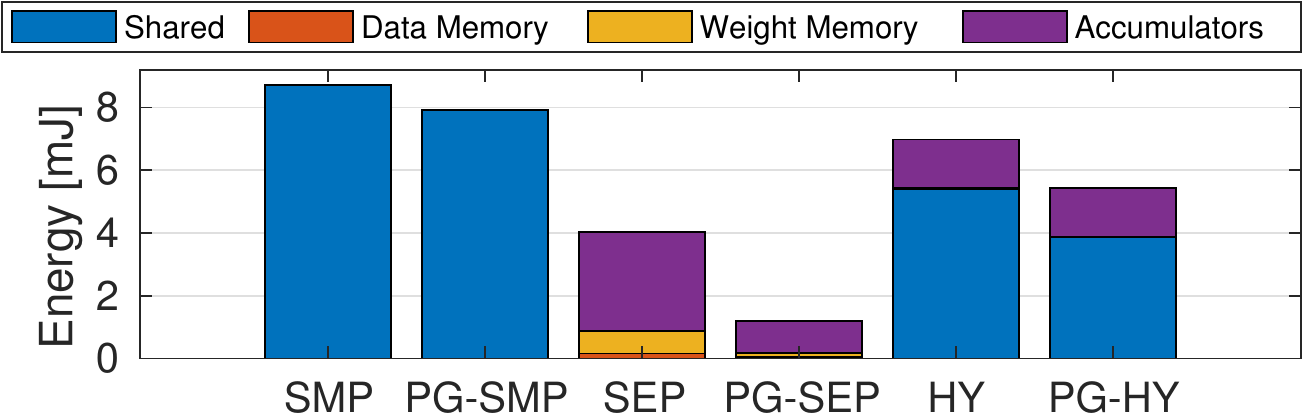}
	\label{fig:energy_versions_components}} \\
	\vspace*{-2mm}
	\subfloat[]{
	\includegraphics[width=\linewidth]{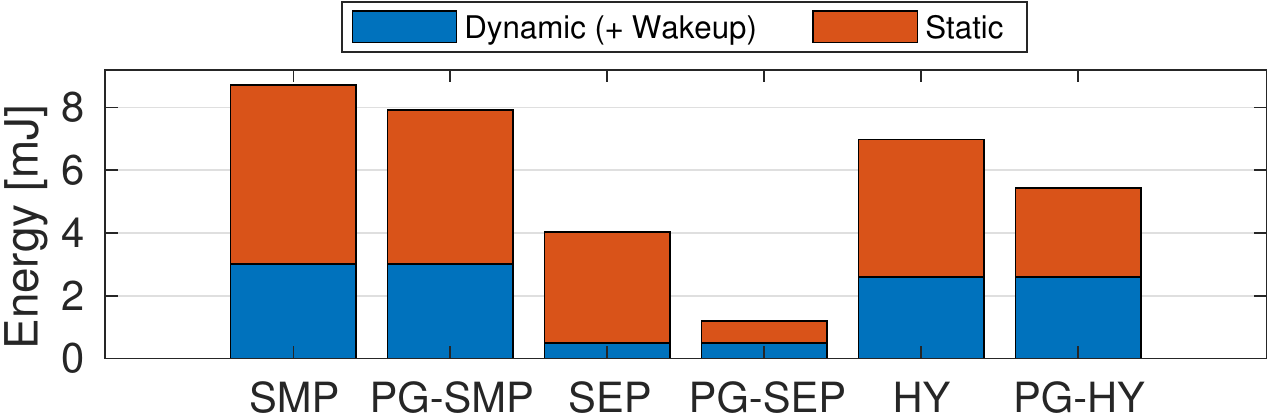}
	\label{fig:energy_versions_dyn_static}} \\
	\vspace*{-2mm}
	\subfloat[]{
	\includegraphics[width=\linewidth]{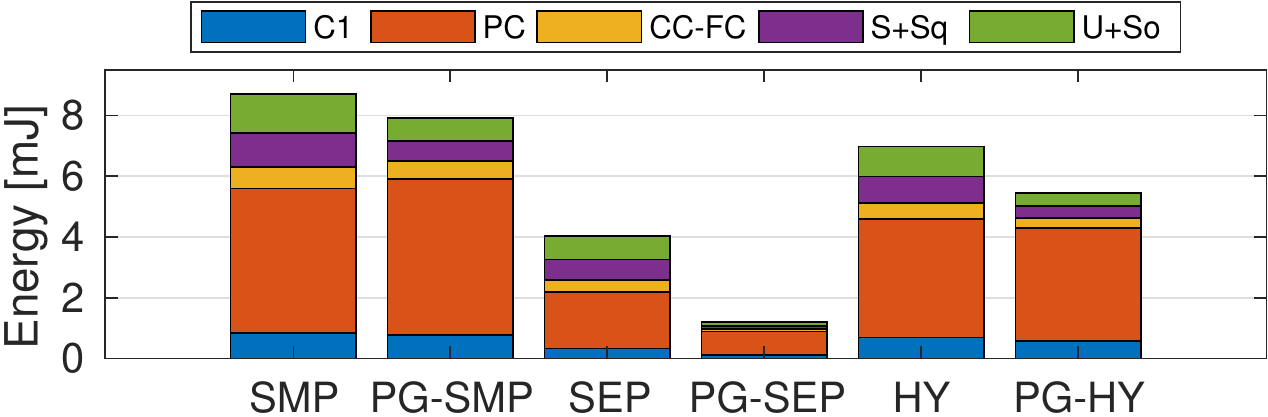}
	\label{fig:energy_versions_operations}}
	\caption{Area and energy results for the different architecture organizations of the CapStore on-chip memory. (a) Area and (b) energy consumption for different memory components. (c) Energy consumption, showing dynamic and static contributions. (d) Energy consumption, for different operations of the CapsuleNet Inference.}
	\label{fig:results}
	\vspace*{0mm}
\end{figure}

For each memory architecture organization explored in \Cref{sec:design}, we evaluate the area and the energy consumption (see \Cref{tab:results_final}), using CACTI-P \cite{ref:CACTI-P} simulation tool. \Cref{fig:area_versions} presents the area breakdown of the different memory components of the CapStore on-chip memory. We notice that, while the organizations SEP and PG-SEP have higher memory size, compared to the other four architectures, the area occupied is significantly lower. This effect is due to having single-port memories instead of shared multi-ports. A shared 3-port memory, indeed, occupies a high area due to the overhead of the interconnections. Moreover, the power-gating circuitry (mainly based on the sleep transistors) is significant in terms of area, as each sleep transistor size depends upon the number of memory cells controlled by itself.

In the following paragraph, we show the energy consumption for different architectures of the CapStore on-chip memory, analyzed from different perspectives. \Cref{fig:energy_versions_components} presents the energy breakdown of the different memory components of the on-chip memory. It is evident that the architectures SEP and PG-SEP are more energy efficient than the others, due to having single-ports. Moreover, the energy consumption can be reduced by applying power-gating. The advantage of using such technique is more significant for the SEP architecture. The reason of this behavior can be explained by considering that the ratio between the \textit{ON} sectors and the complete memory is higher, as compared to SMP and HY.
\Cref{fig:energy_versions_dyn_static} shows the contributions of the dynamic and static energy, for the different architectures. It highlights that (1) \textit{moving from SMP to SEP, we are able to significantly reduce the dynamic energy} and (2) \textit{moving from SEP to PG-SEP, we can significantly reduce the static energy}. Besides this, we noticed that the wakeup energy overhead is negligible, because the transitions of the memory sectors between \textit{OFF} and \textit{ON} are very less frequent (they can only happen when we switch from an operation to the following one), as compared to the periods of time when the states are stable. \Cref{fig:energy_versions_operations} shows the energy breakdown for the different components of the CapsuleNet inference: the proportions of the energy consumed remains approximately similar across the different architectures. We notice that our memory consumes the highest portion of energy for the PrimaryCaps (PC) layer, since it is has intense memory requirements and accesses. Indeed, for this operation, the energy consumed is higher when we do not enable power-gating, since almost every sectors of the memory must be active (the utilization is high), and thus the power-gating is not beneficial.

\vspace*{-2mm}
\subsection{Energy and Area of our Complete CapsuleNet Accelerator Architecture}
\label{subsec:results_complete}

\begin{figure}[t]
\centering
\begin{minipage}[t]{.26\linewidth}
\vspace*{0mm}
\subfloat[]{
\includegraphics[width=\linewidth]{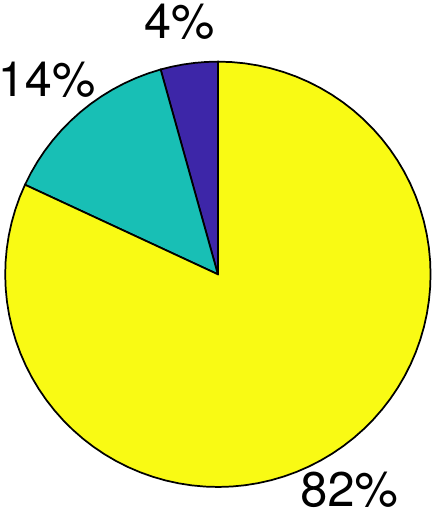}
\label{fig:energy_breakdown_final}}
\end{minipage}
\hfill
\begin{minipage}[t]{.69\linewidth}
\vspace*{0mm}
\subfloat[]{
\includegraphics[width=\linewidth]{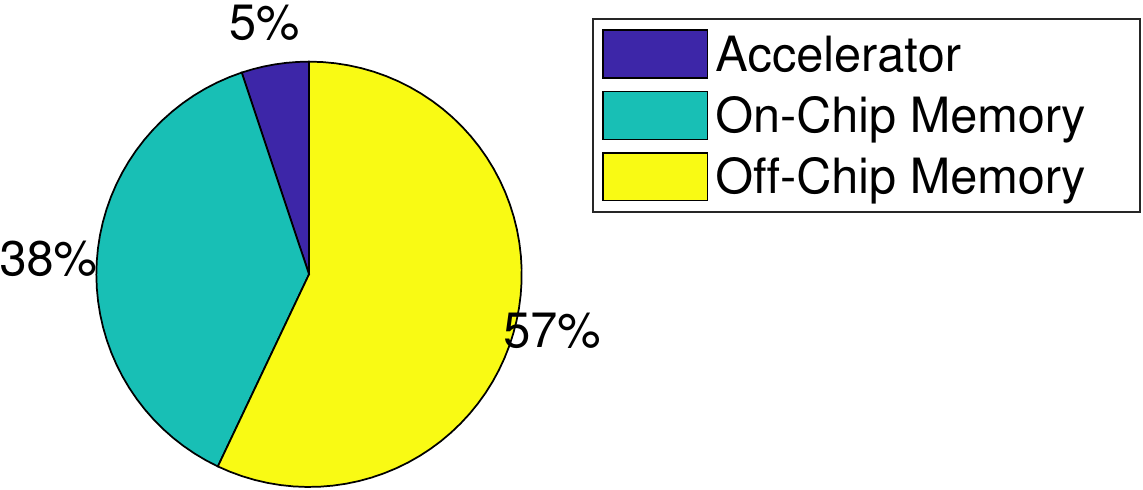}
\label{fig:area_breakdown_final}}
\end{minipage}
\vspace*{0mm}
\caption{(a) Energy and (b) area breakdown of the complete architecture of the accelerator.}
\label{fig:energy_area_breakdown}
\vspace*{0mm}
\end{figure}

Based on the evaluations achieved in \Cref{subsec:results_onchip}, we select the CapStore PG-SEP architecture, as it is the most efficient organization in terms of energy consumption, among the six architectures proposed in the exploration. We synthesize the complete architecture of the CapsuleNet accelerator in a 32nm CMOS technology library, using the ASIC design flow with the Synopsys Design Complier, and we measure the area and energy consumption. \Cref{fig:energy_area_breakdown} shows the energy and the area breakdowns. For both metrics, the contribution of the accelerator to the total value is very limited (4 to 5\%), while the main contribution is due to the off-chip memory. Compared to the initial version, discussed in \Cref{subsec:energy_breakdown} version (a), the total energy is reduced by 78\% and the area by 25\%. Compared to version (b), the on-chip energy is reduced by 86\%, the on-chip area by 47\%. As a consequence, the total energy is reduced by 46\% and the total area by 25\%.

\vspace*{-2mm}
\section{Conclusions}
\label{sec:conclusions}

In this work, after an initial analysis showing the performance and hardware requirements for CapsuleNet inference, we identified that a significant amount of energy can be saved by designing a specialized memory hierarchy (combination of off-chip and on-chip memories). To achieve high efficiency, we designed the on-chip memory in a way to minimize the off-chip memory accesses and maintaining high throughput. We explored different memory hierarchies and we design the CapStore, a specialized on-chip memory for CapsuleNet accelerators, with its own application-aware power management unit to further reduce the leakage power. As per our knowledge, this paper proposes the first on-chip memory design performing the inference on CapsuleNets. Our work opens up interesting ideas for energy-efficient memory design, even beyond CapsuleNet inference applications.

\end{document}